\pdfoutput=1
\documentclass[11pt]{article}

\usepackage[]{acl}

\usepackage{times}
\usepackage{latexsym}

\usepackage[T1]{fontenc}
\usepackage[utf8]{inputenc}

\usepackage{microtype}

\usepackage{inconsolata}

\usepackage{booktabs}
\usepackage{amsmath}
\usepackage{enumitem}
\usepackage{graphicx}
\usepackage{subcaption}
\usepackage{paralist, rotating, multirow}
\usepackage{hyperref}

\newcommand{\instupr}{\textsc{InstUPR }}
\newcommand{\instuprs}{\textsc{InstUPR}}
\title{\instupr: Instruction-based Unsupervised Passage Reranking\\ with Large Language Models}

 \author{Chao-Wei Huang \and Yun-Nung Chen \\
         National Taiwan University \\
         f07922069@csie.ntu.edu.tw \quad y.v.chen@ieee.org}

\begin{document}
\maketitle
\begin{abstract}
This paper introduces \instupr, an unsupervised passage reranking method based on large language models (LLMs).
Different from existing approaches that rely on extensive training with query-document pairs or retrieval-specific instructions, our method leverages the instruction-following capabilities of instruction-tuned LLMs for passage reranking without any additional fine-tuning.
To achieve this, we introduce a soft score aggregation technique and employ pairwise reranking for unsupervised passage reranking.
Experiments on the BEIR benchmark demonstrate that \instupr outperforms unsupervised baselines as well as an instruction-tuned reranker, highlighting its effectiveness and superiority.\footnote{Source code to reproduce all experiments is open-sourced at \url{https://github.com/MiuLab/InstUPR}}

\end{abstract}

\section{Introduction}
Information retrieval (IR) involves the retrieval of relevant information from a large collection of data, such as web pages or documents, in response to a user's query. 
Recently, deep learning methods like dense passage retriever (DPR)~\cite{karpukhin-etal-2020-dense} have gained significant interest due to their superior performance compared to sparse retrieval methods such as BM25. However, it is crucial for initial retrievers to be lightweight to handle a large set of retrieval targets. Therefore, passage reranking plays a crucial role in the process by following the initial retrievers and ranking the retrieved passages based on their relevance to the query. This enables the use of computationally intensive models, thereby enhancing retrieval accuracy.

Large language models (LLMs) have demonstrated strong zero-shot capabilities across various natural language tasks~\cite{brown2020language,kojima2022large}. Specifically, models fine-tuned on natural language instructions have shown remarkable performance in comprehending complex instructions~\cite{weifinetuned}. Previous work has explored the use of LLMs for passage reranking by fine-tuning them on extensive retrieval supervision~\cite{nogueira-etal-2020-document,asai2022tart}. Another line of investigation involves unsupervised passage reranking using LLMs~\cite{sachan-etal-2022-improving,sun2023chatgpt}. However, these unsupervised methods often lack guidance in understanding the relevance of retrieved passages.

This paper introduces \instupr, an instruction-based unsupervised passage reranking method that leverages the instruction-following capabilities of LLMs for reranking \emph{without} the need for labeled relevance information and additional fine-tuning. We employ an instruction-tuned LLM to generate a relevance score for each query-passage pair. Additionally, we propose a soft relevance score aggregation technique that combines the LLM's predicted distribution over possible scores, resulting in robust estimation. We evaluate our method on common evaluation benchmarks, including TREC DL19~\cite{craswell2020overview}, DL20~\cite{craswell2021overview}, and BEIR~\cite{thakur2021beir}. Experimental results demonstrate that our \instupr outperforms unsupervised baselines like UPR and an instruction-tuned reranker. Furthermore, our proposed soft aggregation method significantly contributes to these improvements.

Our contribution can be summarized as 3-fold:
\begin{compactitem}
\item We propose \instupr, which leverages the instruction-following capabilities of LLMs for unsupervised passage reranking.
\item We introduce soft relevance score aggregation to enhance reranking performance.
\item We propose both pointwise and pairwise reranking schemes and demonstrate their effectiveness compared with unsupervised baselines and models specifically fine-tuned on retrieval datasets.
% \item Experiments on popular benchmarks demonstrate the effectiveness of our \instupr compared with unsupervised baselines and models specifically fine-tuned on retrieval datasets.
\end{compactitem}

\begin{figure*}[ht]
    \centering
    \includegraphics[width=\textwidth]{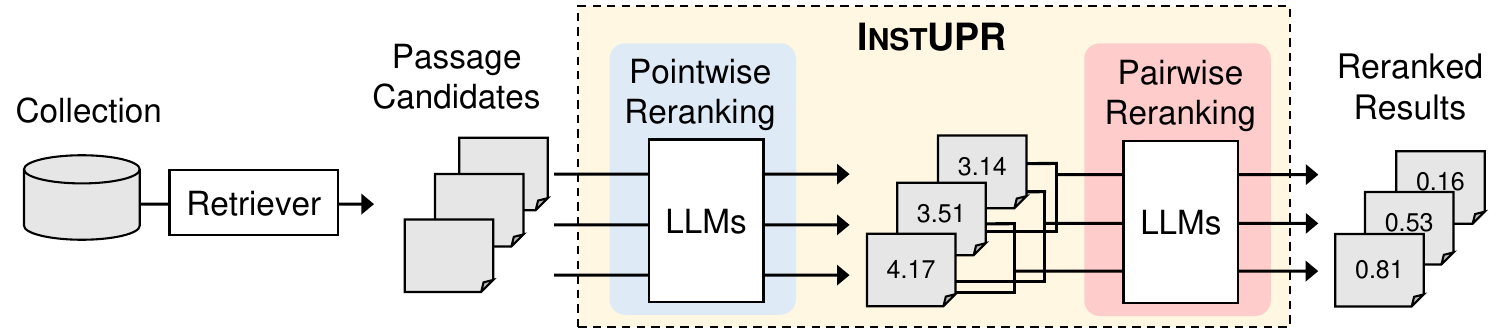}
    \caption{Illustration of our proposed \instupr framework, which includes pointwise reranking and pairwise reranking modules for fine-grained estimation.}
    \label{fig:framework}
\end{figure*}

\section{Related Work}
\paragraph{Information Retrieval}
In recent years, deep learning-based retrieval models have achieved remarkable performance across various information retrieval tasks. The dense passage retriever (DPR) framework, which encodes documents and queries into dense representations, has emerged as a popular approach for dense retrieval~\cite{karpukhin-etal-2020-dense}. With the advent of large language models (LLMs), numerous methods have leveraged these models for dense retrieval. GTR~\cite{ni-etal-2022-large} utilizes LLM encoders for dense retrieval and demonstrates performance improvements with increased model size. Promptagator~\cite{dai2023promptagator} and InPars~\cite{bonifacio2022inpars} propose the use of LLMs to generate synthetic query-document pairs, which are then employed for training dense retrievers. 
Our work is orthogonal to these methods, as we focus on utilizing LLMs for second-stage passage reranking.

\paragraph{Passage Reranking}
Passage reranking typically serves as a second-stage component following large-scale retrieval. Several studies have proposed deep reranking models that encode query-document pairs to predict relevance scores~\cite{nogueira2019passage}. 
\citet{nogueira-etal-2020-document} introduced a generation-based method for passage reranking by fine-tuning LLMs on MS-MARCO\cite{bajaj2016ms}, a large-scale retrieval dataset with relevance annotations. Their model, MonoT5, generates the word \texttt{true} for relevant pairs and \texttt{false} for irrelevant pairs. Similarly, our method also adopts a generation-based approach. The main difference is that our method does not require relevance annotations nor fine-tuning; instead, we leverage the instruction-following capabilities of LLMs to enable unsupervised estimation. TART~\cite{asai2022tart} fine-tunes LLMs on extensive retrieval supervision from various tasks with instructions. Our method differs from TART in that we do not require any retrieval supervision and employ a generation-based approach in an unsupervised fashion.

Another research line is unsupervised passage reranking with LLMs, which eliminates the need for retrieval supervision. UPR~\cite{sachan-etal-2022-improving} is the pioneering attempt at unsupervised passage reranking, proposing to rerank passages by estimating the conditional likelihood of generating the query given the passage using LLMs. UPR has shown promising results, but it employs an indirect measure that may not be optimal for measuring the relevance of retrieved passages. In contrast, our \instupr leverages the instruction-following capabilities of LLMs while requiring no retrieval supervision. Through extensive experiments, we demonstrate that \instupr outperforms UPR on most datasets, highlighting its effectiveness. Concurrent to our work\footnote{At the time of writing. This manuscript was originally written and submitted in June 2023.}, \citet{sun2023chatgpt} and \citet{ma2023zero} both proposed to perform listwise passage reranking by prompting ChatGPT, which is a black-box commercial system\footnote{\url{https://chat.openai.com/}}. Our work focuses on pointwise and pairwise reranking, and employs an open-sourced LLM with well-documented data sources to facilitate scientific understanding of our method.

\section{Our Method}
The task of passage reranking involves assigning a relevance score to each document in a set of retrieved candidates given a query. Formally, given a query $q$ and a set of retrieved passages $D = {d_1, d_2, \cdots, d_k}$, a reranker aims to assign a relevance score to each query-passage pair as $s(q, d_i)$. These relevance scores are then used to rerank the passage candidates. Figure~\ref{fig:framework} illustrates the proposed reranking framework.

\subsection{\instuprs: Instruction-based Unsupervised Passage Reranking}
\begin{figure}[t]
    \centering
    \begin{subfigure}{0.45\textwidth}
        \includegraphics[width=\textwidth]{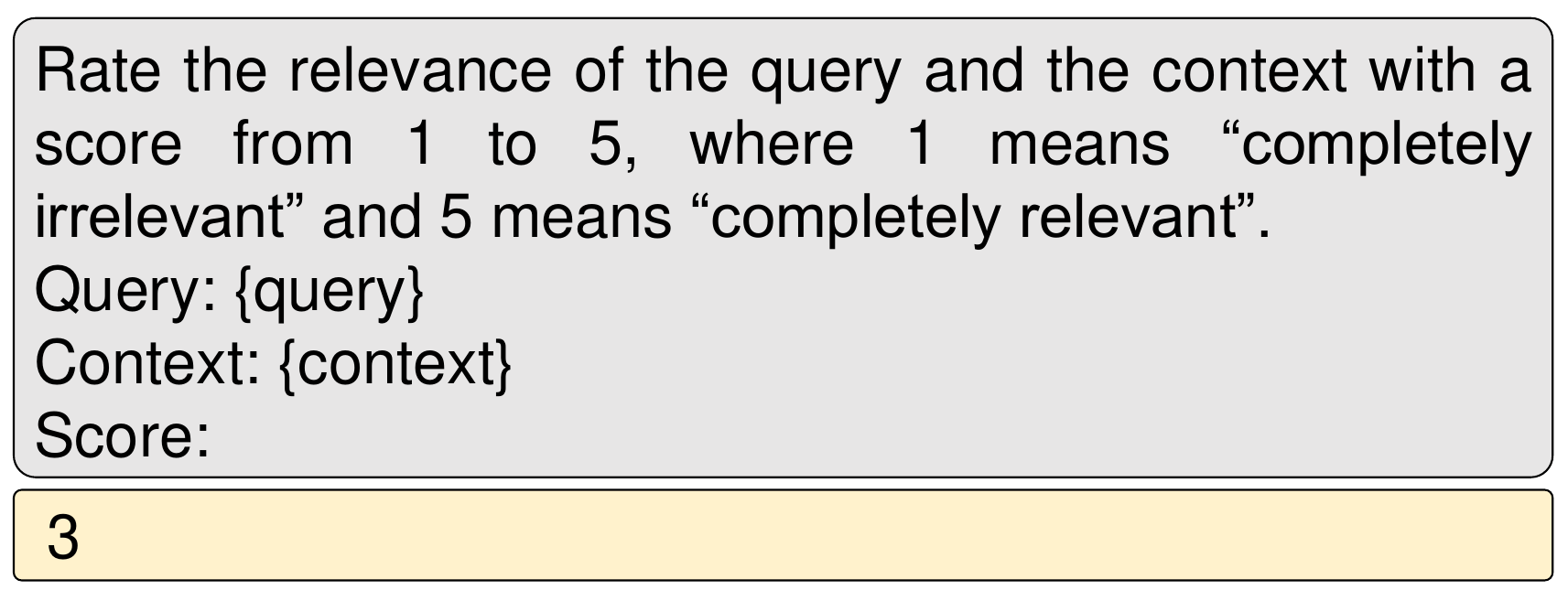}
        \caption{Instrcution for pointwise reranking.}
        \label{fig:instruction_pointwise}
    \end{subfigure}
    \hfill
    \begin{subfigure}{0.45\textwidth}
        \includegraphics[width=\textwidth]{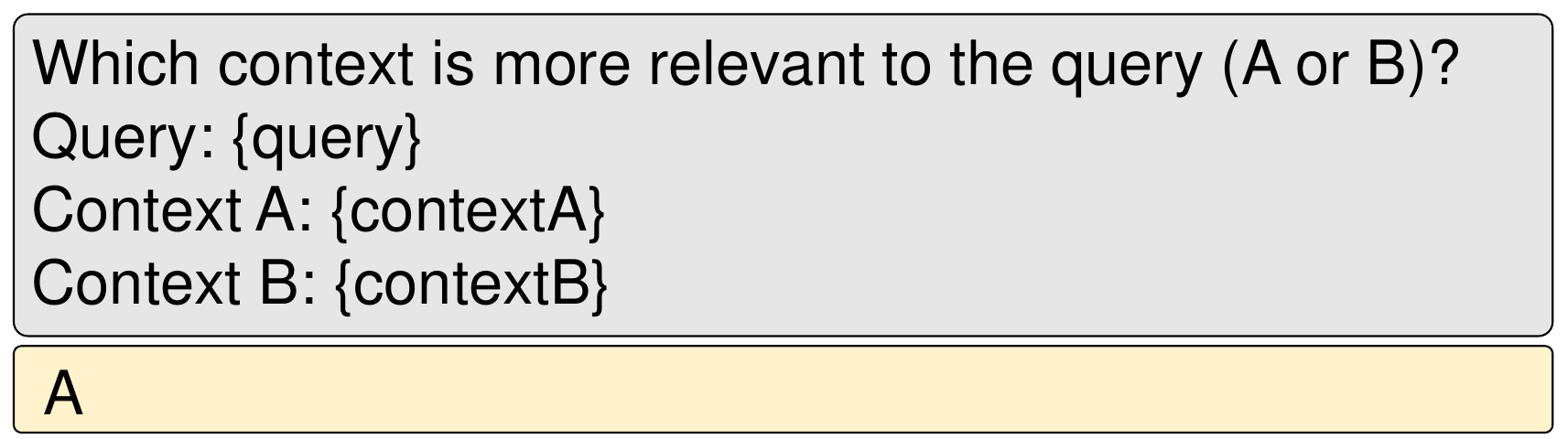}
        \caption{Instruction for pairwise reranking.}
        \label{fig:instruction_pairwise}
    \end{subfigure}
    \hfill
    \caption{The instruction templates for reranking in \instuprs.}
    \label{fig:instructions}
\end{figure}

Our method, \instuprs, leverages the instruction-following capabilities of LLMs to enhance the performance of passage reranking. We prompt the LLMs with task-specific instructions that instruct them to directly generate a relevance score for each query-passage pair $(q, d_i)$ and rerank the passage candidates based on their relevance scores. In this paper, we instruct the LLMs to predict a relevance score from 1 to 5 using the Likert scale. For parsing convenience, we instruct the LLMs to generate only a single token from the options, which in our case are ${1, 2, 3, 4, 5}$. An example of the instruction template is shown in Figure~\ref{fig:instruction_pointwise}.

\subsection{Soft Relevance Score Aggregation}
Generating a single token as the relevance score introduces several issues~\cite{liu2023gpteval}. 
First, it results in discrete scores that lead to many ties, which is suboptimal for reranking.
Second, we observe that the generated scores tend to be very similar for the same task, such as the LLM frequently outputting a score of 3 for the majority of the passages. 
To address these issues, we propose \textit{Soft Relevance Score Aggregation}. Instead of using the generated token directly, we compute a weighted sum of the options using their probabilities as weights. Specifically, the soft relevance score of a query-passage pair $s_1(q, d_i)$ can be calculated as:
\begin{equation*}
s_1(q, d_i) = \sum_{n=1}^5 n \cdot p(n \mid q, d_i),
\end{equation*}
where $p(n \mid q, d_i)$ is the probability of predicting a score of $n$ by the LLM. This score can also be interpreted as the expected value of the score predicted by the LLM.

\subsection{Pairwise Reranking}
Pairwise reranking has been demonstrated to outperform pointwise reranking while being more computationally expensive~\cite{nogueira2019multi,pradeep2021expando}. Given a query $q$ and two passages $d_i$ and $d_j$, we instruct the LLM to select the passage that is more relevant to the query and assign the probability of selecting each passage as the score. The final score of a passage $d_i$, denoted as $s_2(q, d_i)$, is then re-estimated as the sum of its scores against all other passage candidates:
\begin{equation*}
s_2(q, d_i) = \sum_{i \neq j} p(i \mid q, d_i, d_j),
\end{equation*}
where $p(i \mid q, d_i, d_j)$ is the probability predicted by the LLM of $d_i$ being more relevant to the query $q$ than $d_j$. It is important to note that the ordering of passages affects the scores, i.e., $p(i \mid q, d_i, d_j) \neq p(i \mid q, d_j, d_i)$. Therefore, we evaluate all $(k^2 - k)$ pairs to obtain the pairwise rankings for robustness. An instruction template is shown in Figure~\ref{fig:instruction_pairwise}.

\begin{table*}
\centering
\resizebox{.9\linewidth}{!}{%
\begin{tabular}{c|lc|cc|ccc}
\toprule
& & & \multicolumn{2}{c|}{\textbf{Supervised}} & \multicolumn{3}{c}{\textbf{Unsupervised}} \\
& & \bf BM25 & \bf TART-Rerank & \bf MonoT5-3B & \bf UPR & \bf \instuprs$_{point}$ & \bf \instuprs$_{+pair}$  \\
\midrule
\multirow{2}{*}{\rotatebox{90}{TREC}} & DL19 & 50.58 & 67.43 & \bf 71.83 & 54.51 & 61.61 & \underline{70.53} \\
& DL20 & 47.96 & 59.19 & \bf 68.89 & 55.91 & 61.88 & \underline{68.55} \\
\midrule
%\multicolumn{7}{l}{\bf BEIR} \\
%\midrule
\multirow{13}{*}{\rotatebox{90}{BEIR}} &
TREC-COVID & 59.47 & 74.20 & \underline{80.71} & 69.25 & 73.04 & \bf 81.33  \\
& BioASQ & 52.25 & 56.20 & \underline{57.50} & 56.59 & 55.17 & \bf 59.25  \\
& NFCorpus & 32.18 & 33.70 & \bf 38.97 & 33.78 & 35.24 & \underline{37.10} \\
& FiQA & 23.61 & 35.70 & \bf 45.99 & 37.19 & 39.76 & \underline{41.24}  \\
& Signal-1M & \bf 33.04 & 28.38 & 32.55 & 31.78 & \underline{32.58} & 31.26 \\
& TREC-News & 39.52 & 42.63 & \underline{48.49} & 36.06 & 46.12 & \bf 50.37 \\
& Robust04 & 40.70 & 50.63 & \underline{56.71} & 44.40 & 54.03 & \bf 60.23 \\
& Touche-2020 & \bf 44.22 & 28.33 & 32.41 & 21.07 & 28.98 & \underline{34.22} \\
& DBPedia & 31.80 & 42.53 & \bf 44.45 & 30.72 & 42.43 & \underline{43.53} \\
& SCIDOCS & 14.90 & 17.34 & \underline{19.00} & 15.88 & 18.97 & \bf 20.68 \\
& Climate-FEVER & 16.51 & \underline{27.21} & \bf 27.33 & 18.25 & 26.18 & 25.81 \\
& SciFact & 64.76 & \underline{75.19} & \bf 76.57 & 73.09 & 71.28 & 74.96 \\
\cmidrule{2-8}
& Average (BEIR) & 38.01 & 42.67 & \bf 46.65 & 39.01 & 43.66 & \underline{46.63} \\
\bottomrule
\end{tabular}
}
\caption{Reranking performance (NDCG@10) of both supervised and unsupervised methods (\%); the best scores are in \textbf{bold}, and the second best scores are \underline{underlined}.}
\label{tab:results}
\end{table*}

\section{Experiments}
\subsection{Setup}
To evaluate the effectiveness of our proposed \instuprs, we conduct experiments on TREC DL19~\cite{craswell2020overview}, DL20~\cite{craswell2021overview}, and BEIR~\cite{thakur2021beir}, which consists of various tasks for zero-shot retrieval and reranking.
Following previous work, we employ BM25 as the base retrieval method and retrieve the top-100 passages for reranking~\cite{rosa2022no}.
For our experiments, we utilize \texttt{flan-t5-xl}~\cite{https://doi.org/10.48550/arxiv.2210.11416} as our LLM to ensure that it has not been pretrained on our specific datasets.
We report \textbf{NDCG@10}, which is the standard metric for evaluating retrieval performance.
Additional details can be found in Appendix~\ref{sec:appendix_details}.

\subsection{Baseline Systems}
\begin{compactitem}
    \item \textbf{TART-Rerank}~\cite{asai2022tart} is a state-of-the-art reranker that is fine-tuned on a collection of retrieval datasets with instructions.
    \item \textbf{MonoT5-3B}~\cite{nogueira2019passage} is a reranker that is fine-tuned on MS MARCO for predicting whether the passage is relevant to the query.
    \item \textbf{UPR}~\cite{sachan-etal-2022-improving} is an unsupervised reranking method that reranks passages by their conditional probabilities of generating the query. We use \texttt{flan-t5-xl} for UPR for a fair comparison.
\end{compactitem}

\begin{table}[ht]
\centering
\begin{tabular}{lccc}
\toprule
 & DL19 & DL20 & BEIR \\
\midrule
\instuprs$_{point}$ & \bf 61.61 & \bf 61.88 & \bf 43.66 \\
 - soft aggregation & 57.08 & 58.13 & 37.13 \\
 - Likert scale & 58.27 & 57.46 & 38.36 \\
\bottomrule
\end{tabular}
\caption{Results of ablation study (\%).}
\label{tab:ablation}
\end{table}

%\section{Results}
\subsection{Main Results}
The experimental results are presented in Table~\ref{tab:results}. 
In comparison to the unsupervised baseline UPR, our \instuprs$_{point}$ outperforms UPR in 12 out of the 14 tasks, exhibiting an average relative improvement of over 10\%. Furthermore, \instuprs$_{point}$ outperforms TART-Rerank in 8 tasks, despite not being trained with any retrieval supervision.
It highlights the effectiveness of our proposed instruction-based reranking method, which directly leverages the instruction-following capabilities of LLMs.

With the inclusion of our proposed unsupervised pairwise reranking (\instuprs$_{+pair}$), we achieve the best performance in 5 tasks and the second-best performance in 6 tasks. 
Remarkably, \instuprs$_{+pair}$ achieves comparable performance to the state-of-the-art reranker MonoT5-3B while being an unsupervised method, demonstrating its practical value for real-world applications.

\subsection{Ablation Study}
To validate the effectiveness of individual components, we conduct an ablation study presented in Table~\ref{tab:ablation}. 
Removing the soft score aggregation component leads to significant degradation in all tasks, highlighting the importance of our proposed soft score aggregation for robust estimation. 
We also examine the impact of removing the Likert scale and directly asking the LLM whether the passage is relevant to the query, using the probability of generating ``\texttt{yes}'' as the relevance score. 
The results demonstrate a substantial drop after removing the Likert-based scores, showing the effectiveness of our proposed scoring method. 
% Given the challenge of controlling LLMs, the presence of a robust and reasonable scoring method is crucial, and our experiments validate the efficacy of our proposed approach.

\section{Conclusion}
In this paper, we propose \instuprs, an instruction-based unsupervised passage reranking method.
We leverage the instruction-following capabilities of LLMs for passage reranking and propose soft score aggregation and pairwise reranking to further improve the performance.
Experimental results show that \instuprs outperforms previous unsupervised methods and achieves comparable performance to the state-of-the-art method, demonstrating the great potential of leveraging LLMs for information retrieval tasks.
We hope our work can draw attention to exploring the application of LLMs to information retrieval studies.
Future work could explore how the scale of LLMs affects reranking performance and efficient pairwise reranking techniques. % , as well as how improved reranking performance could benefit dense retrieval.

\section*{Limitations}
While our proposed method demonstrates impressive performance, it is important to acknowledge certain limitations. 
First, the pairwise reranking approach we employ incurs high computational costs, making it challenging to scale up to scenarios involving hundreds of passage candidates. Future research could focus on exploring more efficient pairwise reranking techniques to address this limitation.
Second, our experiments are conducted using a single large language model (LLM), and it is possible that different LLMs may exhibit varying behaviors and performances. To address this, further investigation is needed to assess the generalization capabilities across diverse LLMs.

\section*{Ethics Statement}
In this study, we utilize an instruction-following LLM that has been pre-trained on extensive text data and subsequent fine-tuning with instructions. It is important to recognize that LLMs have the potential to exhibit biased and offensive behavior, which can impact the quality and veracity of the reranking results. Careful attention should be given to mitigating bias and ensuring ethical considerations are taken into account when deploying such models in real-world applications.

\section*{Acknowledgements}
This work was financially supported by the National Science and Technology Council (NSTC) in Taiwan, under Grants 111-2222-E-002-013-MY3, 111-2628-E-002-016, and 112-2223-E002-012-MY5.

\bibliography{anthology,custom}

\begin{thebibliography}{22}
\expandafter\ifx\csname natexlab\endcsname\relax\def\natexlab#1{#1}\fi

\bibitem[{Asai et~al.(2022)Asai, Schick, Lewis, Chen, Izacard, Riedel,
  Hajishirzi, and Yih}]{asai2022tart}
Akari Asai, Timo Schick, Patrick Lewis, Xilun Chen, Gautier Izacard, Sebastian
  Riedel, Hannaneh Hajishirzi, and Wen-tau Yih. 2022.
\newblock Task-aware retrieval with instructions.
\newblock \emph{arXiv preprint arXiv:2211.09260}.

\bibitem[{Bajaj et~al.(2016)Bajaj, Campos, Craswell, Deng, Gao, Liu, Majumder,
  McNamara, Mitra, Nguyen et~al.}]{bajaj2016ms}
Payal Bajaj, Daniel Campos, Nick Craswell, Li~Deng, Jianfeng Gao, Xiaodong Liu,
  Rangan Majumder, Andrew McNamara, Bhaskar Mitra, Tri Nguyen, et~al. 2016.
\newblock Ms marco: A human generated machine reading comprehension dataset.
\newblock \emph{arXiv preprint arXiv:1611.09268}.

\bibitem[{Bonifacio et~al.(2022)Bonifacio, Abonizio, Fadaee, and
  Nogueira}]{bonifacio2022inpars}
Luiz Bonifacio, Hugo Abonizio, Marzieh Fadaee, and Rodrigo Nogueira. 2022.
\newblock Inpars: Unsupervised dataset generation for information retrieval.
\newblock In \emph{Proceedings of the 45th International ACM SIGIR Conference
  on Research and Development in Information Retrieval}, pages 2387--2392.

\bibitem[{Brown et~al.(2020)Brown, Mann, Ryder, Subbiah, Kaplan, Dhariwal,
  Neelakantan, Shyam, Sastry, Askell et~al.}]{brown2020language}
Tom Brown, Benjamin Mann, Nick Ryder, Melanie Subbiah, Jared~D Kaplan, Prafulla
  Dhariwal, Arvind Neelakantan, Pranav Shyam, Girish Sastry, Amanda Askell,
  et~al. 2020.
\newblock Language models are few-shot learners.
\newblock \emph{Advances in neural information processing systems},
  33:1877--1901.

\bibitem[{Chung et~al.(2022)Chung, Hou, Longpre, Zoph, Tay, Fedus, Li, Wang,
  Dehghani, Brahma, Webson, Gu, Dai, Suzgun, Chen, Chowdhery, Narang, Mishra,
  Yu, Zhao, Huang, Dai, Yu, Petrov, Chi, Dean, Devlin, Roberts, Zhou, Le, and
  Wei}]{https://doi.org/10.48550/arxiv.2210.11416}
Hyung~Won Chung, Le~Hou, Shayne Longpre, Barret Zoph, Yi~Tay, William Fedus,
  Eric Li, Xuezhi Wang, Mostafa Dehghani, Siddhartha Brahma, Albert Webson,
  Shixiang~Shane Gu, Zhuyun Dai, Mirac Suzgun, Xinyun Chen, Aakanksha
  Chowdhery, Sharan Narang, Gaurav Mishra, Adams Yu, Vincent Zhao, Yanping
  Huang, Andrew Dai, Hongkun Yu, Slav Petrov, Ed~H. Chi, Jeff Dean, Jacob
  Devlin, Adam Roberts, Denny Zhou, Quoc~V. Le, and Jason Wei. 2022.
\newblock \href {https://doi.org/10.48550/ARXIV.2210.11416} {Scaling
  instruction-finetuned language models}.

\bibitem[{Craswell et~al.(2021)Craswell, Mitra, Yilmaz, and
  Campos}]{craswell2021overview}
Nick Craswell, Bhaskar Mitra, Emine Yilmaz, and Daniel Campos. 2021.
\newblock Overview of the trec 2020 deep learning track. corr abs/2102.07662
  (2021).
\newblock \emph{arXiv preprint arXiv:2102.07662}.

\bibitem[{Craswell et~al.(2020)Craswell, Mitra, Yilmaz, Campos, and
  Voorhees}]{craswell2020overview}
Nick Craswell, Bhaskar Mitra, Emine Yilmaz, Daniel Campos, and Ellen~M
  Voorhees. 2020.
\newblock Overview of the trec 2019 deep learning track.
\newblock \emph{arXiv preprint arXiv:2003.07820}.

\bibitem[{Dai et~al.(2023)Dai, Zhao, Ma, Luan, Ni, Lu, Bakalov, Guu, Hall, and
  Chang}]{dai2023promptagator}
Zhuyun Dai, Vincent~Y Zhao, Ji~Ma, Yi~Luan, Jianmo Ni, Jing Lu, Anton Bakalov,
  Kelvin Guu, Keith Hall, and Ming-Wei Chang. 2023.
\newblock \href {https://openreview.net/forum?id=gmL46YMpu2J} {Promptagator:
  Few-shot dense retrieval from 8 examples}.
\newblock In \emph{The Eleventh International Conference on Learning
  Representations}.

\bibitem[{Karpukhin et~al.(2020)Karpukhin, Oguz, Min, Lewis, Wu, Edunov, Chen,
  and Yih}]{karpukhin-etal-2020-dense}
Vladimir Karpukhin, Barlas Oguz, Sewon Min, Patrick Lewis, Ledell Wu, Sergey
  Edunov, Danqi Chen, and Wen-tau Yih. 2020.
\newblock \href {https://doi.org/10.18653/v1/2020.emnlp-main.550} {Dense
  passage retrieval for open-domain question answering}.
\newblock In \emph{Proceedings of the 2020 Conference on Empirical Methods in
  Natural Language Processing (EMNLP)}, pages 6769--6781, Online. Association
  for Computational Linguistics.

\bibitem[{Kojima et~al.(2022)Kojima, Gu, Reid, Matsuo, and
  Iwasawa}]{kojima2022large}
Takeshi Kojima, Shixiang~Shane Gu, Machel Reid, Yutaka Matsuo, and Yusuke
  Iwasawa. 2022.
\newblock Large language models are zero-shot reasoners.
\newblock In \emph{ICML 2022 Workshop on Knowledge Retrieval and Language
  Models}.

\bibitem[{Liu et~al.(2023)Liu, Iter, Xu, Wang, Xu, and Zhu}]{liu2023gpteval}
Yang Liu, Dan Iter, Yichong Xu, Shuohang Wang, Ruochen Xu, and Chenguang Zhu.
  2023.
\newblock Gpteval: Nlg evaluation using gpt-4 with better human alignment.
\newblock \emph{arXiv preprint arXiv:2303.16634}.

\bibitem[{Ma et~al.(2023)Ma, Zhang, Pradeep, and Lin}]{ma2023zero}
Xueguang Ma, Xinyu Zhang, Ronak Pradeep, and Jimmy Lin. 2023.
\newblock Zero-shot listwise document reranking with a large language model.
\newblock \emph{arXiv preprint arXiv:2305.02156}.

\bibitem[{Ni et~al.(2022)Ni, Qu, Lu, Dai, Hernandez~Abrego, Ma, Zhao, Luan,
  Hall, Chang, and Yang}]{ni-etal-2022-large}
Jianmo Ni, Chen Qu, Jing Lu, Zhuyun Dai, Gustavo Hernandez~Abrego, Ji~Ma,
  Vincent Zhao, Yi~Luan, Keith Hall, Ming-Wei Chang, and Yinfei Yang. 2022.
\newblock \href {https://aclanthology.org/2022.emnlp-main.669} {Large dual
  encoders are generalizable retrievers}.
\newblock In \emph{Proceedings of the 2022 Conference on Empirical Methods in
  Natural Language Processing}, pages 9844--9855, Abu Dhabi, United Arab
  Emirates. Association for Computational Linguistics.

\bibitem[{Nogueira and Cho(2019)}]{nogueira2019passage}
Rodrigo Nogueira and Kyunghyun Cho. 2019.
\newblock Passage re-ranking with bert.
\newblock \emph{arXiv preprint arXiv:1901.04085}.

\bibitem[{Nogueira et~al.(2020)Nogueira, Jiang, Pradeep, and
  Lin}]{nogueira-etal-2020-document}
Rodrigo Nogueira, Zhiying Jiang, Ronak Pradeep, and Jimmy Lin. 2020.
\newblock \href {https://doi.org/10.18653/v1/2020.findings-emnlp.63} {Document
  ranking with a pretrained sequence-to-sequence model}.
\newblock In \emph{Findings of the Association for Computational Linguistics:
  EMNLP 2020}, pages 708--718, Online. Association for Computational
  Linguistics.

\bibitem[{Nogueira et~al.(2019)Nogueira, Yang, Cho, and
  Lin}]{nogueira2019multi}
Rodrigo Nogueira, Wei Yang, Kyunghyun Cho, and Jimmy Lin. 2019.
\newblock Multi-stage document ranking with bert.
\newblock \emph{arXiv preprint arXiv:1910.14424}.

\bibitem[{Pradeep et~al.(2021)Pradeep, Nogueira, and Lin}]{pradeep2021expando}
Ronak Pradeep, Rodrigo Nogueira, and Jimmy Lin. 2021.
\newblock The expando-mono-duo design pattern for text ranking with pretrained
  sequence-to-sequence models.
\newblock \emph{arXiv preprint arXiv:2101.05667}.

\bibitem[{Rosa et~al.(2022)Rosa, Bonifacio, Jeronymo, Abonizio, Fadaee, Lotufo,
  and Nogueira}]{rosa2022no}
Guilherme~Moraes Rosa, Luiz Bonifacio, Vitor Jeronymo, Hugo Abonizio, Marzieh
  Fadaee, Roberto Lotufo, and Rodrigo Nogueira. 2022.
\newblock No parameter left behind: How distillation and model size affect
  zero-shot retrieval.
\newblock \emph{arXiv preprint arXiv:2206.02873}.

\bibitem[{Sachan et~al.(2022)Sachan, Lewis, Joshi, Aghajanyan, Yih, Pineau, and
  Zettlemoyer}]{sachan-etal-2022-improving}
Devendra Sachan, Mike Lewis, Mandar Joshi, Armen Aghajanyan, Wen-tau Yih,
  Joelle Pineau, and Luke Zettlemoyer. 2022.
\newblock \href {https://aclanthology.org/2022.emnlp-main.249} {Improving
  passage retrieval with zero-shot question generation}.
\newblock In \emph{Proceedings of the 2022 Conference on Empirical Methods in
  Natural Language Processing}, pages 3781--3797, Abu Dhabi, United Arab
  Emirates. Association for Computational Linguistics.

\bibitem[{Sun et~al.(2023)Sun, Yan, Ma, Ren, Yin, and Ren}]{sun2023chatgpt}
Weiwei Sun, Lingyong Yan, Xinyu Ma, Pengjie Ren, Dawei Yin, and Zhaochun Ren.
  2023.
\newblock Is chatgpt good at search? investigating large language models as
  re-ranking agent.
\newblock \emph{arXiv preprint arXiv:2304.09542}.

\bibitem[{Thakur et~al.(2021)Thakur, Reimers, R{\"u}ckl{\'e}, Srivastava, and
  Gurevych}]{thakur2021beir}
Nandan Thakur, Nils Reimers, Andreas R{\"u}ckl{\'e}, Abhishek Srivastava, and
  Iryna Gurevych. 2021.
\newblock \href {https://openreview.net/forum?id=wCu6T5xFjeJ} {{BEIR}: A
  heterogeneous benchmark for zero-shot evaluation of information retrieval
  models}.
\newblock In \emph{Thirty-fifth Conference on Neural Information Processing
  Systems Datasets and Benchmarks Track (Round 2)}.

\bibitem[{Wei et~al.(2021)Wei, Bosma, Zhao, Guu, Yu, Lester, Du, Dai, and
  Le}]{weifinetuned}
Jason Wei, Maarten Bosma, Vincent Zhao, Kelvin Guu, Adams~Wei Yu, Brian Lester,
  Nan Du, Andrew~M Dai, and Quoc~V Le. 2021.
\newblock Finetuned language models are zero-shot learners.
\newblock In \emph{International Conference on Learning Representations}.

\end{thebibliography}

\appendix
\section{Additional Details}
\label{sec:appendix_details}
\subsection{Dataset}
For fair comparisons, we exclude the datasets NaturalQuestions, HotpotQA, Quora, and FEVER from BEIR as they are part of either our LLM's training set or the baselines' training set. Additionally, we exclude CQADupStack due to its evaluation complexity and its large number of queries. Also, we exclude Arguana since it is a passage-level retrieval task.

\subsection{Implementation Details}
For pairwise reranking, given the top-k retrieval results, we evaluate $(k^2 - k)$ pairs to obtain the pairwise scores.
To reduce computations, we reduce $k$ from 100 to 40 for smaller datasets and 15 for larger datasets.
All experiments are conducted on 2xNVIDIA V100 GPUs.
Future work could explore efficient pairwise reranking algorithms, such as applying sorting algorithms to pairwise reranking.

\end{document}